\documentclass[10pt,conference]{IEEEtran}
\IEEEoverridecommandlockouts

\usepackage{cite}

\usepackage{color}

\ifCLASSINFOpdf
   \usepackage[pdftex]{graphicx}
   \graphicspath{{figs/}}
   \DeclareGraphicsExtensions{.pdf,.jpeg,.png}
\else
   \usepackage[dvips]{graphicx}
   \graphicspath{{../figs/}}
   \DeclareGraphicsExtensions{.eps}
\fi
\usepackage[cmex10]{amsmath}
\usepackage{amsthm}

\usepackage{algorithmic}

\usepackage{array}

\usepackage{multicol}

\ifCLASSOPTIONcompsoc
  \usepackage[caption=false,font=normalsize,labelfont=sf,textfont=sf]{subfig}
\else
  \usepackage[caption=false,font=footnotesize]{subfig}
\fi
\usepackage{url}

\begin{document}
%
\title{On the Learning of Deep Local Features for\\Robust Face Spoofing Detection}

\newif\iffinal
\finaltrue
\newcommand{\jemsid}{99999}


\iffinal

\author{\IEEEauthorblockN{Gustavo Botelho de Souza$^1$, Jo\~{a}o Paulo Papa$^2$ and Aparecido Nilceu Marana$^2$  - \textbf{{\color{blue}in Proc. of SIBGRAPI 2018}}}
\IEEEauthorblockA{
$^1$UFSCar - Federal University of S\~{a}o Carlos. Rod. Washington Lu\'{i}s, Km 235. S\~{a}o Carlos (SP), Brazil. 13565-905.\\
$^2$UNESP - S\~{a}o Paulo State University. Av. Eng. Luiz Edmundo Carrijo Coube, 14-01. Bauru (SP), Brazil. 17033-360.\\
E-mail: gustavo.botelho@gmail.com, \{papa, nilceu\}@fc.unesp.br}
}


%

\else
  \author{Authors\\Institutions\\SIBGRAPI paper ID: \jemsid \\ }
\fi

\maketitle

\begin{abstract}
Biometrics emerged as a robust solution for security systems. However, given the dissemination of biometric applications, criminals are developing techniques to circumvent them by simulating physical or behavioral traits of legal users (spoofing attacks). Despite face being a promising characteristic due to its universality, acceptability and presence of cameras almost everywhere, face recognition systems are extremely vulnerable to such frauds since they can be easily fooled with common printed facial photographs. State-of-the-art approaches, based on Convolutional Neural Networks (CNNs), present good results in face spoofing detection. However, these methods do not consider the importance of learning deep local features from each facial region, even though it is known from face recognition that each facial region presents different visual aspects, which can also be exploited for face spoofing detection. In this work we propose a novel CNN architecture trained in two steps for such task. Initially, each part of the neural network learns features from a given facial region. Afterwards, the whole model is fine-tuned on the whole facial images. Results show that such pre-training step allows the CNN to learn different local spoofing cues, improving the performance and the convergence speed of the final model, outperforming the state-of-the-art approaches.

\end{abstract}


\IEEEpeerreviewmaketitle

\section{Introduction}

Biometric systems are increasingly common in our everyday activities \cite{bib:Jain1}. People recognition through their own physical, physiological or behavioral traits inhibits most of the frauds often committed in security systems based on knowledge (passwords) or tokens (cards, keys,  etc.). However, nowadays criminals are already developing techniques to accurately simulate the biometric characteristics of valid users, such as face, fingerprint and iris, to gain access to places or systems, process known as spoofing attack \cite{bib:Patel, bib:Ratha}. In this context, robust countermeasure techniques must be developed and integrated into the traditional biometric applications in order to prevent such frauds. 
Despite face being a promising trait due to its convenience for users, universality and acceptability, traditional face recognition systems can be easily fooled with common printed facial photographs~\cite{bib:Patel}, which nowadays can be obtained by criminals on the worldwide network, especially due to the dissemination of social medias and networks.

Spatial image information
is extremely important in tasks involving faces, such as face detection \cite{bib:ViolaJones} and face recognition \cite{bib:Eigenfaces, bib:Chiachia}. The different visual patterns of each facial region
encode rich and discriminative information necessary to distinguish a face from other objects, and also 
from other faces. Regarding face spoofing detection,
some works based on handcrafted features have mentioned that different spoofing cues can be extracted from different facial regions \cite{bib:Ak,bib:Chingovska}.

Recently, deep learning architectures have emerged as good alternatives for solving complex problems and have reached state-of-the-art results in many tasks due to their great power of abstraction and robustness, working with high-level features, self-learned from the training data \cite{bib:Bengio,analysisCNN}. Among the proposed deep learning architectures, Convolutional Neural Networks (CNN)~\cite{bib:LeCun} have appeared as one of the most important classes of deep neural networks able to deal with digital images with great performances. 

Some CNN based state-of-the-art methods were recently proposed for face spoofing detection \cite{bib:Atoum,  bib:Li, bib:Xu, bib:ref3}.
However, none of them take into account the different visual aspects of each facial region and, consequently, the different local spoofing cues that could be learned by the neural networks to improve their performances. All proposed methods work on whole faces, in a holistic way, or with random and small patches, i.e., they train the neural networks with samples extracted from random regions of the faces, all together. This can degrade the performance of the training algorithm since the backpropagation method can be distracted by the different visual information
extracted from random regions of the face, instead of 
learning
the real differences between real and fake faces in each facial region, with similar visual aspects, differing only by spoofing cues.

In this context, we propose a novel CNN architecture trained in two steps for a better performance in face spoofing detection: (i) the local pre-training phase, in which each part of the model is trained on each main facial region, learning deep local features for attack detection and initializing the whole model in a great position in the search space (the network learns to detect multiple and different spoofing cues from all the facial regions); (ii) the global fine tuning phase, in which the whole model is fine-tuned based on the weights learned independently by its parts and on whole real and fake facial images, in order to improve the model generalization. Results obtained on two major datasets used for the evaluation of face spoofing detection techniques, Replay-Attack \cite{bib:Chingovska} and CASIA FASD (Face Antispoofing Database) \cite{bib:Zhang}, show that the pre-training step on local and fixed regions of the faces improves the performance of the final model and its convergence speed. The proposed approach outperformed the state-of-the-art methods while working with an efficient CNN architecture.

\section{Technical Background}

In this section we briefly present some concepts regarding the importance of spatial information and differences of the facial regions for face detection, face recognition and face spoofing detection, as well as some related works.

\subsection{Facial Regions and Spatial Information}

The spatial relationship between the facial elements and regions in the images encodes rich information that can be used to distinguish a face from the background, from other objects or even from other faces \cite{bib:ViolaJones, bib:Eigenfaces}. The first works on automated face detection and recognition already used such kind of information, presenting good results and efficiency.

\begin{figure}[h]
\centering
\includegraphics[width=0.37\textwidth]{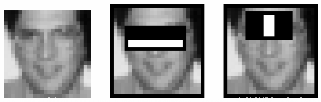}
\caption{Left: face detected based on the Haar-like features used by Viola and Jones \cite{bib:ViolaJones}. Center and right: examples of Haar-like features. Images from the  OpenCV documentation \cite{bib:OpenCV}.}
\label{fig:viola}
\end{figure}

Regarding face detection, the early work of Viola and Jones~\cite{bib:ViolaJones} used Haar-like features to detect the presence of faces in digital images. In short, they apply, to each area of a given image, a cascade classifier which verifies, hierarchically, whether all main facial features are present in that area. The Haar-like features are designed to capture typical differences existing in neighboring regions of human faces. Fig. \ref{fig:viola} shows two Haar-like features and their correspondence to the regions of human faces. The black rectangles indicate that darker regions are expected, while white rectangles indicate that brighter regions are expected in a certain area. The feature showed in the middle focus on darker and brighter regions corresponding to the eyes (especially due to eyebrows) and cheeks, respectively. The feature on the right searches for the contrast of the nose and eyes in human faces.

Based on the work of Viola and Jones~\cite{bib:ViolaJones}, which allowed automated face detection in reasonable time for real applications, many works were later proposed that also explored the contrasts existing in neighboring regions of the face~\cite{bib:FaceDetect1, bib:FaceDetect2, bib:FaceDetect3}. 

In the context of face recognition, the first effective method for real scenarios
was proposed by Turk and Pentland~\cite{bib:Eigenfaces}, based on the Principal Component Analysis (PCA)~\cite{bib:PCA}, which can be used to find the most discriminative eigenvectors that best describe the variance of the set of data under analysis (facial images, in this case)
and reduce the dimensionality of the problem. 
Given the similarity of such eigenvectors (when represented as 2D images) to facial images, 
Turk and Pentland called them eigenfaces \cite{bib:Eigenfaces, bib:FigEig}. 
It is possible to identify the facial elements and regions (and their spatial relationship) in the eigenfaces, indicating that this kind of information is important to differentiate faces from different people.



Works based on other transformations for reducing the dimensionality of the facial images space, such as the ones based on the Linear Discriminat Analysis (LDA) \cite{bib:Fisher}, also usually obtain, as the ``basis" of the new coordinate systems, vectors that ensemble human faces when viewed as 2D images, with the different facial regions in them being noticeable. The CNN based architectures for face recognition, which self-learn the most discriminative features for face representation from the training datasets, also capture the spatial information and relationships between facial elements and regions, presenting connection weights between neurons that act as edge and facial elements detectors (eyes, nose, etc.)~\cite{vggface}.

Researches in Psychology
show that human beings have an extreme ability to detect faces, more accurately and much faster than any other object, and also highlighted the importance of spatial information and the positioning of each facial region and element for face detection and recognition \cite{bib:Psic1,bib:Psic2}. In \cite{bib:Psic2}, for instance, the authors found that the time required by a group of people to identify a visual stimulus as a face was shorter when normal faces were presented than when jumbled faces, i.e., faces with parts out of place (the mouth region above the eyes, etc.), were presented. 

Despite all this, to the best of our knowledge, no work has investigated the usage of deep local features, learned from each facial region (with its particular visual aspect), to improve the performance of the state-of-the-art deep learning architectures for face spoofing detection, our main goal.

\subsection{Face Spoofing Detection}




According to Ratha, Connell and Bolle \cite{bib:Ratha}, as in any other security system, there are many ways to attack a biometric system. In short, the attacks to biometric applications can be divided in two groups: direct and indirect attacks. In the direct attacks (spoofing attacks), criminals generate synthetic samples of biometric traits of legal users, such as photographs (face simulation), gelatin fingers (fingerprint simulation), contact lenses (iris simulation), among others, to obtain access to places or systems. Criminals try to fool the capture sensor with such samples, the most vulnerable point of the biometric recognition system \cite{bib:Ratha}. 

\begin{figure}[h]
\centering
\includegraphics[width=0.49\textwidth]{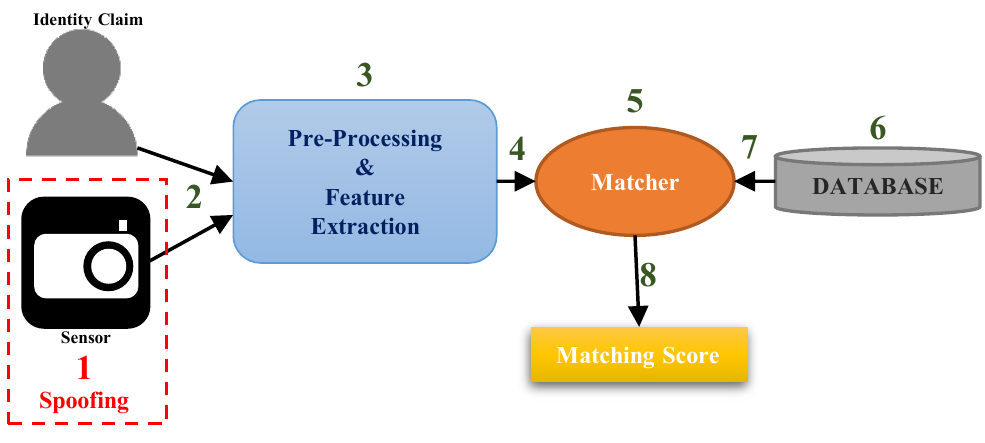}
\caption{Points of attack in a traditional biometric system. The spoofing attacks occur in point ``1", i.e., by fooling the sensor (presentation of fake traits) \cite{bib:Galbally}.}
\label{fig:Ataques}
\end{figure}

In the indirect attacks, criminals, after investigating the inner working of the system and based on some fragility, act by modifying the algorithms used to match templates or internal messages exchanged by the system modules \cite{bib:Ratha}. Fig.~\ref{fig:Ataques} shows the main points of attack of a biometric system. It is important to know, however, that the vast majority of attacks on biometric applications are direct, due to the simplicity for attackers who do not need to investigate the inner working of the system. 

\begin{figure}[b]
\centering
\includegraphics[width=0.49\textwidth]{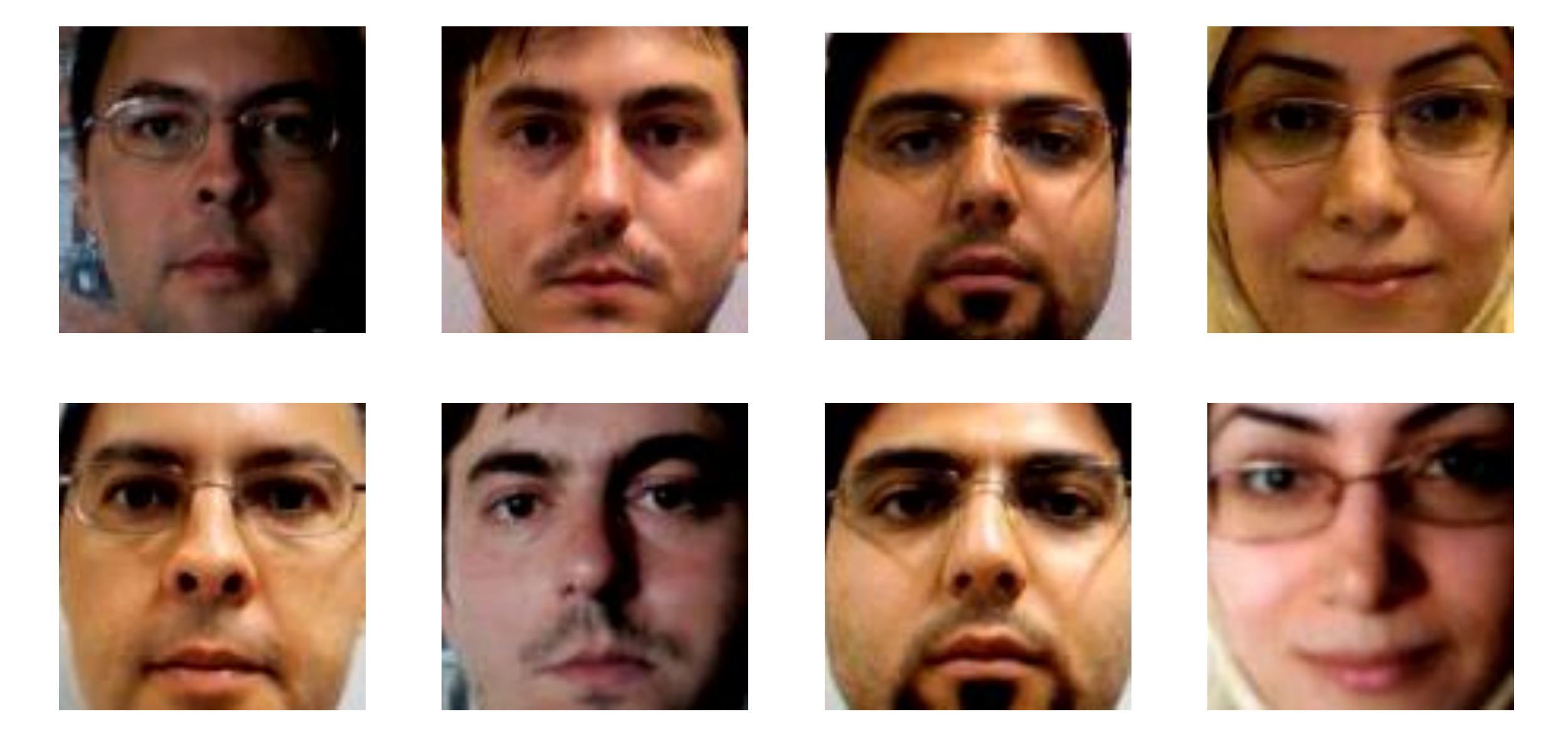}
\caption{Images from real (first row) and fake faces (second row) from the Replay-Attack \cite{bib:Chingovska} dataset. 
}
\label{fig:Faces}
\end{figure}

Among the main biometric traits, as said, face is a promising one especially due to its convenience, low cost of acquisition, universality and acceptability by users \cite{bib:Jain1}, being very suitable to a wide variety of environments, including mobile ones. However, despite all these advantages, face recognition systems are the ones that most suffer from spoofing attacks since they can be easily fooled even with common printed photographs \cite{bib:Patel}. Fig.~\ref{fig:Faces} shows some real and fake faces from the Replay-Attack \cite{bib:Chingovska} dataset. As one can observe, it is very difficult to distinguish between real and fake faces.

Regarding face spoofing attacks, these can be performed by presenting to the camera of a biometric system a static face image (printed, digital image displayed on a mobile device, or a 3D mask) or a dynamic set of face images (videos recorded from the faces of legal users displayed on mobile devices)~\cite{bib:Patel}. As one can observe in Fig.~\ref{fig:Faces}, different spoofing cues can be analyzed in each facial region, such as shadows (more common in real faces, especially in their outer regions, than in 2D fake faces).

\subsection{Related Works}

Face spoofing detection methods have been proposed in literature in the last years. Regarding the approaches that work with handcrafted features, most of them focus on detecting spoofing artifacts and image quality distortions in order to identify fake faces. Some of them, such as \cite{bib:Quality1}, extract color features, based on the assumption that, when recaptured by the cameras of the biometric systems, fake faces present distortions in colors, reflectance, etc., due to the properties of the materials they are made with. In \cite{bib:Quality1}, the authors argue that fake faces tend to present darker colors and different contrasts, as well as more low-frequency areas than real faces. They use such information to extract features for face classification.

Other works  extract texture features based on the LBP (Local Binary Patterns)\cite{bib:LBP}  descriptor and its variations, to characterize real and fake faces, presenting good results
\cite{bib:MLBP, bib:LBPTOP,bib:DLTP}. In \cite{bib:MLBP}, the authors extract specific features from each facial region and combine them into a final feature vector in the end of the process for classification, improving significantly the results of the method compared to when working with features from the whole face. Some of these works also mentioned that the best features were extracted from specific facial regions, especially the central one and from the forehead area \cite{bib:MLBP}. 

Among the approaches for face spoofing detection which use deep learning architectures, more specifically Convolutional Neural Networks (CNN) \cite{bib:LeCun}, since for this task they obtained the state-of-the-art results, to the best of our knowledge, all of them work on whole faces, learning global spoofing cues, or on random and small patches extracted from the faces, not focusing on the learning of local spoofing cues from each facial region. In \cite{bib:Li}, for instance, the authors apply a transfer learning algorithm in order to adapt the VGG-Face~\cite{vggface} model, a benchmark CNN for face recognition trained on $2.6$ million facial images from $2,622$ people, for spoofing detection, obtaining good results given the similar domains of the problems. In \cite{bib:CIARP2017}, the authors also apply a similar transfer learning algorithm on VGG-Face~\cite{vggface} model using it for feature extraction, without modifying the original model, focusing on efficiency. 
In \cite{bib:Lotufo}, a more time consuming algorithm for transfer learning is applied to the VGG-Face~\cite{vggface} neural network, in which layers of the original CNN are updated for the spoofing detection task, obtaining great results, but also making the process more expensive and requiring more processing power (advanced GPUs) and time. 

All these aforementioned works based on the VGG-Face~\cite{vggface} consider whole facial images as input. Other important works in the literature such as \cite{bib:ref3, bib:UnicampAll} also extract global deep spoofing cues from the faces based on other architectures. In \cite{bib:Xu}, the authors propose a CNN model and integrate it with a Long-Short Term Memory (LSTM) \cite{bib:LSTM} neural network for learning temporal holistic features from the faces in sequences of images (videos), also obtaining a good performance.    

In \cite{bib:Atoum}, the authors explore random patches for face spoofing detection. They use such approach especially for augmenting the dataset but present the patches all together (from random and different parts of the faces) to train their CNN architecture. Despite the good results, given the different visual patterns of each facial region, the neural network can be distracted and base its learning for spoofing detection
mainly on the structural information of the faces,
much more evident in the images, 
not focusing on the spoofing cues themselves.
In other words, the backpropagation algorithm
can be more influenced by the structural aspects of the facial elements (e.g. presence or absence, size, shape, etc. of the eyes) in a given patch, than by the subtle spoofing cues in it.

Another well-known patch based approach for face spoofing detection, presented in \cite{bib:Ak}, works with small and not fixed patches (regions) from the faces to train traditional classification models. In each face, given an extensive analysis based on several metrics, they select the best patches to represent the whole facial image in order to classify it as real or fake. They use many metrics to determine which patches should be selected to represent the face, which are obtained from different regions of the faces for each sample, also degrading the performance of the method in learning spoofing cues.

Despite the lack of attention to deep local features regarding face spoofing detection, Krizhevsky \cite{bib:Kriz} demonstrated, on other image classification tasks, that the use of local (and fixed) regions of the images (visual local information), in an initial training step of the deep learning model, tends to improve its performance, also avoiding getting stuck in local minima in the hyperparameter search space. Ba et al. \cite{bib:Neuro} also suggested the use of facial patches for initializing deep models applied to face recognition based on studies in Neuroscience. Another work \cite{bib:Daniel} uses this initial training step based on fixed image patches for improving vehicle classification in images.

\section{Proposed Approach}

In this work, we propose a novel CNN architecture for face spoofing detection, which we called lsCNN (Locally Specialized CNN), with a novel training algorithm for a more effective learning of deep local spoofing features, based on two steps: (i) the local pre-training phase, in which each part of the model is trained on each main facial region (predefined and fixed), learning deep local features for attack detection 
and allowing to initialize the whole model in a
better 
position in the search space; and (ii) the global fine tuning phase, in which the whole model is fine-tuned based on the weights learned independently by its parts on the facial regions, in order to improve its generalization.

\subsection{lsCNN Architecture}

Basically, the lsCNN presents 4 convolutional and pooling layers ($Conv1/Pool1$ to $Conv4/Pool4$) at the bottom, with each convolutional layer being immediately followed by a batch normalization, scale and signal rectification (ReLU - Rectified Linear Unit) layers. The batch normalization and scale layers serve to normalize the output feature maps obtained in the convolutional layers, improving learning \cite{bib:BN}. The rectification function, in each neuron, acts as activation function, eliminating negative values in the resultant feature maps and also accelerating training. At the top of the network is a fully-connected layer ($FC1$), also followed by a batch normalization, scale and ReLU layers, as well as a dropout one ($Drop1$). Finally, there is a softmax layer with two neurons in order to classify the faces as being real or fake. Tab. \ref{tab:LSCNN} presents the lsCNN architecture in terms of its layers, i.e., size of kernels, strides, sizes of input and output feature maps. 

\begin{table}[h]
\centering
\caption{Architecture of the proposed lsCNN. The inputs of lsCNN are RGB (3 channels) facial images with $96\times96$ pixels: $3\times(96\times96)$ maps.}
\begin{tabular}{|c|c|c|c|c|} 
\hline
\textbf{Layer} & \textbf{Kernel Size} & \textbf{Stride} & \textbf{Input Maps} & \textbf{Output Maps} \\ 
\hline \hline
Conv1 & $3\times3$         & 1      & $3\times(96\times96)$  & $27\times(94\times94)$   \\ 
\hline
Pool1                       & $2\times2$                              & 2                           & $27\times(94\times94)$                      & $27\times(47\times47)$                        \\\hline
Conv2                       & $3\times3$                              & 1                           & $27\times(47\times47)$                      & $36\times(45\times45)$                        \\\hline
Pool2                       & $2\times2$                              & 2                           & $36\times(45\times45)$                      & $36\times(23\times23)$                        \\\hline
Conv3                       & $3\times3$                              & 1                           & $36\times(23\times23)$                      & $45\times(21\times21)$                        \\\hline
Pool3                       & $2\times2$                              & 2                           & $45\times(21\times21)$                      & $45\times(11\times11)$                        \\\hline
Conv4                       & $3\times3$                              & 1                           & $45\times(11\times11)$                      & $54\times(9\times9)$                          \\\hline
Pool4                       & $2\times2$                              & 2                           & $54\times(9\times9)$                        & $54\times(5\times5)$                          \\\hline
FC1                         & ---                              & ---                         & $54\times(5\times5)$                        & $1\times(450)$                           \\\hline
Drop1                       & ---                              & ---                         & ---                             & ---                               \\\hline
Softmax                     & ---                              & ---                         & $1\times(450)$                         & $1\times(2)$        \\

\hline

\end{tabular}
\label{tab:LSCNN}
\end{table}

As shown in Tab. \ref{tab:LSCNN}, lsCNN expects 3-channels facial images in RGB color space as input. Although other color spaces allow dealing more accurately with illumination issues, in order to approximate the model to the inner working of human eyes (which capture only red, green and blue waves of light) and their perception in natural conditions, as well as by the fact that most digital cameras capture 
images
in RGB mode, we opted for this image representation over other color models.

\subsection{Local Pre-training}

\begin{figure}[h]
\centering
\includegraphics[width=0.27\textwidth]{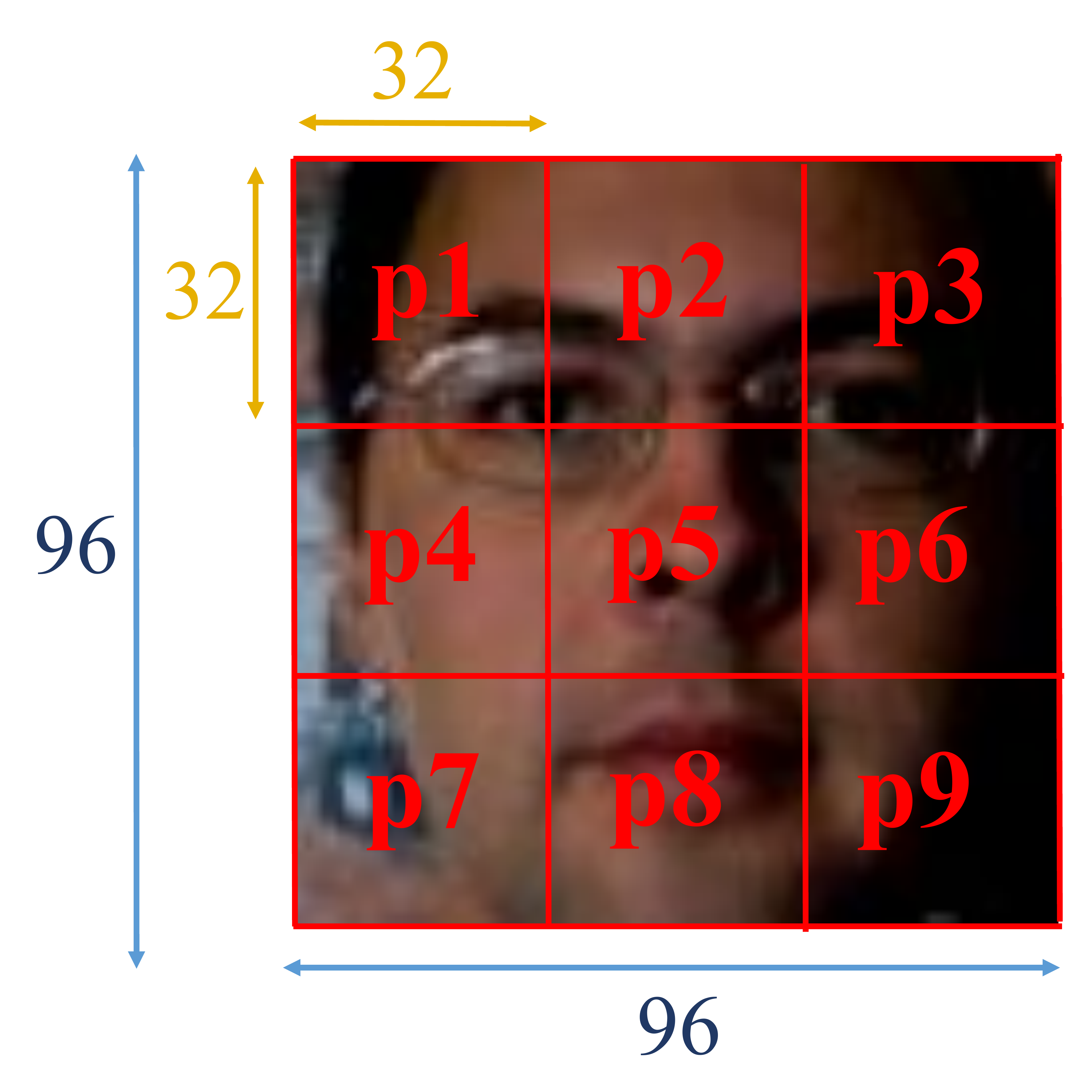}
\caption{A face image ($96\times96$ pixels) from the Replay-Attack  dataset~\cite{bib:Chingovska} split into 9 fixed patches (non-overlapping regions of $32\times32$ pixels).}
\label{fig:Patches}
\end{figure}

Similar to \cite{bib:Kriz}, in order to initialize the whole lsCNN model in a 
better position in the search space and make it specialized in deep local spoofing features from each region of the faces, we split each training face into 9 main regions (patches), as shown in Fig. \ref{fig:Patches}, regions also adopted for face recognition \cite{bib:Jain1}.

\begin{figure}[b]
\centering
\includegraphics[width=0.417\textwidth]{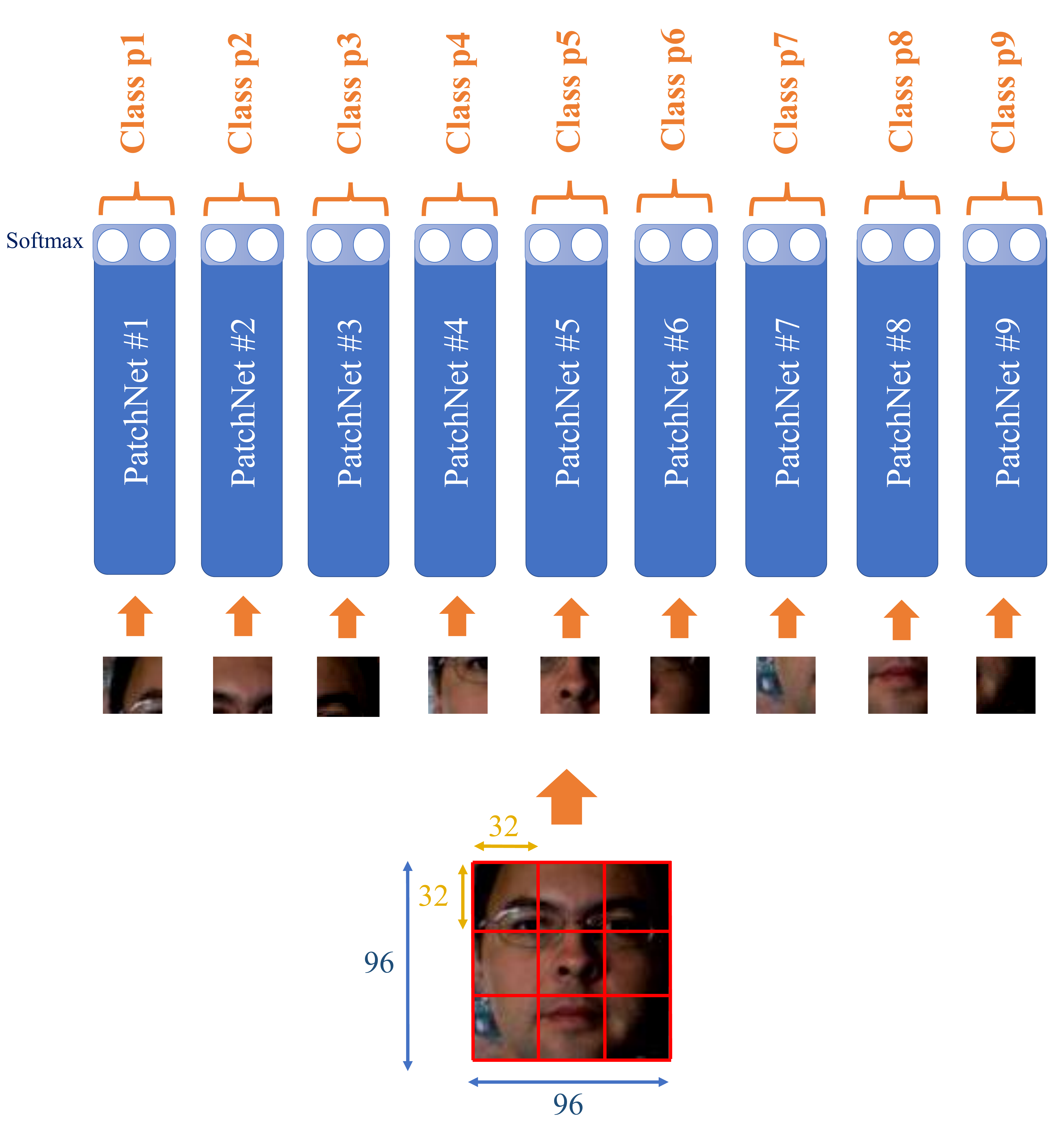}
\caption{Illustration of the local pre-training process of lsCNN. Given a facial image, it is split into its 9 main regions, from $p1$ to $p9$, and 9 instances of the smaller CNN architecture (PatchNet) are trained on each of them.}
\label{fig:Pretraining}
\end{figure}

After this, we also split the lsCNN architecture into 9 independent smaller CNNs, called PatchNets for simplicity, presenting, each of them, a ninth of the size of the original model, and being trained on each of the 9 main facial regions considered from the faces, from $p1$ to $p9$. Each PatchNet has as input RGB patches with $32\times32$ pixels from a respective region of the training faces. Tab. \ref{tab:LSCNNsmall} shows the architecture of each PatchNet and Fig. \ref{fig:Pretraining} illustrates the training process of the 9 instances of this smaller neural network on the facial regions of a given image. As one can observe, on the top of each PatchNet are 2 softmax neurons since they are trained to classify their respective patches as being real or fake.

\begin{table}[h]
\centering
\caption{Architecture of each smaller CNN (PatchNet), part of the lsCNN, trained on each facial region, from $p1$ to $p9$ (fixed patches with $32\times32$ pixels, also in RGB color space).}
\begin{tabular}{|c|c|c|c|c|} 
\hline
\textbf{Layer} & \textbf{Kernel Size} & \textbf{Stride} & \textbf{Input Maps} & \textbf{Output Maps} \\ 
\hline \hline
Conv1 & $3\times3$         & 1      & $3\times(32\times32)$  & $3\times(30\times30)$   \\ 
\hline
Pool1                       & $2\times2$                              & 2                           & $3\times(30\times30)$                      & $3\times(15\times15)$                        \\\hline
Conv2                       & $3\times3$                              & 1                           & $3\times(15\times15)$                      & $4\times(13\times13)$                        \\\hline
Pool2                       & $2\times2$                              & 2                           & $4\times(13\times13)$                      & $4\times(7\times7)$                        \\\hline
Conv3                       & $3\times3$                              & 1                           & $4\times(7\times7)$                      & $5\times(5\times5)$                        \\\hline
Pool3                       & $2\times2$                              & 2                           & $5\times(5\times5)$                      & $5\times(3\times3)$                        \\\hline
Conv4                       & $3\times3$                              & 1                           & $5\times(3\times3)$                      & $6\times(1\times1)$                          \\\hline
Pool4                       & $2\times2$                              & 2                           & $6\times(1\times1)$                        & $6\times(1\times1)$                          \\\hline
FC1                         & ---                              & ---                         & $6\times(1\times1)$                        & $1\times(50)$                           \\\hline
Drop1                       & ---                              & ---                         & ---                             & ---                               \\\hline
Softmax                     & ---                              & ---                         & $1\times(50)$                         & $1\times(2)$        \\

\hline

\end{tabular}
\label{tab:LSCNNsmall}
\end{table}

\subsection{Global Fine Tuning}

After training the 9 smaller neural networks in their respective facial regions, their weights and biases are used to initialize the parts of the whole lsCNN for a fine tuning step of such larger model on the whole training facial images, in order to improve its generalization. 

\begin{figure*}[h]
    \begin{center}
        \includegraphics[width=1\textwidth]{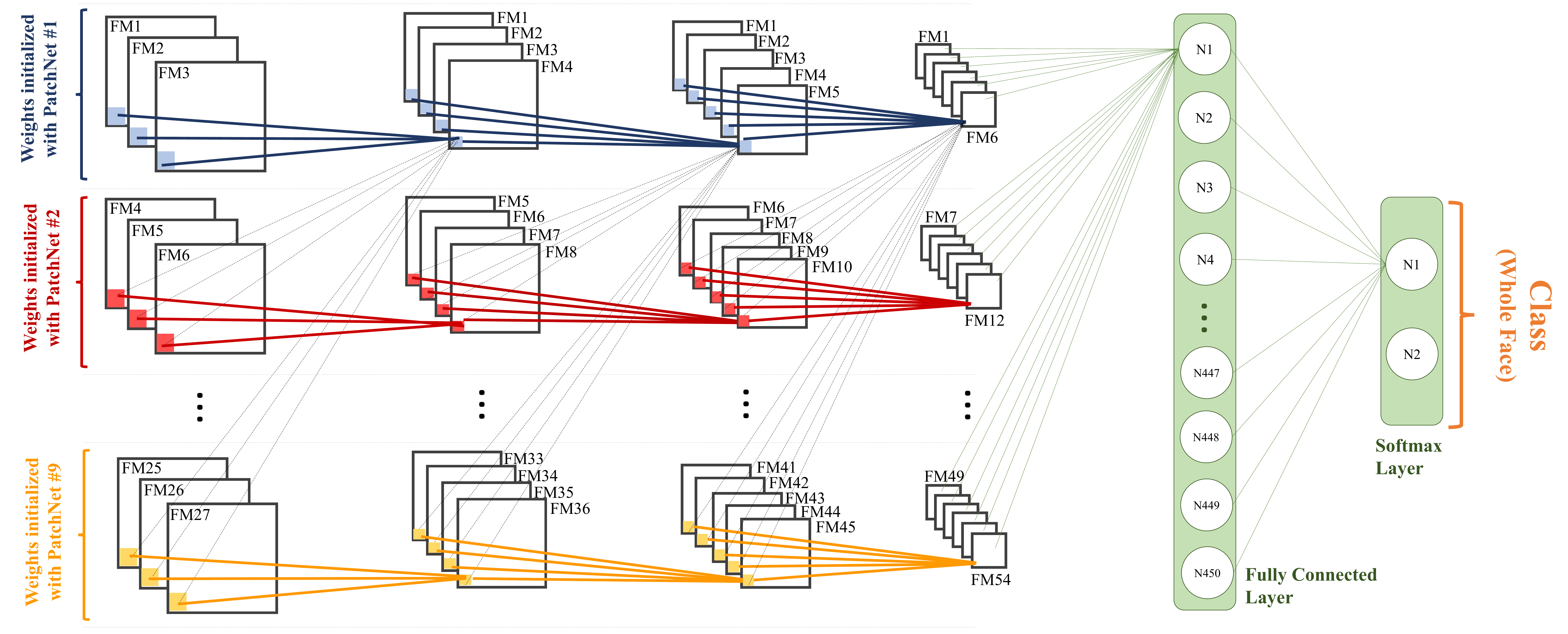}
        \caption{Initialization of the lsCNN  model based on the weights of the 9 PatchNets. The thicker colored lines represent $3\times3$ connections and are initialized with the weights learned by each PatchNet. The first PatchNet, for example, initializes the weights between the first neurons (first feature maps - FM) in the layers of lsCNN. The thin black dotted lines also indicate $3\times3$ connections zero-initialized and the green thin ones are initialized with random values from a normal distribution (zero-mean and standard deviation of $0.01$, by default).  The thin gray lines are just for a better visualization of the initialization process.}
        \label{fig:initialization}
    \end{center}
\end{figure*}

As shown in Fig. \ref{fig:initialization}, each smaller network initializes the weights of the connections and biases of a partition (a ninth) of the lsCNN model, from the left (top) to the right (bottom) side of the lsCNN model. The weights of the first PatchNet, for example, initialize the connections between the most left neurons of the lsCNN model, responsible for first feature maps ($FM1$ to $FM3$ in the case of the first network layer), and so on (similarly to \cite{bib:Kriz}). The connections of lsCNN between neurons from different parts of it  are zero-initialized. 

The weights of the two fully-connected layers on top are randomly initialized from a normal distribution in order to improve the generalization of model even more, as in \cite{bib:Kriz}. Their biases are zero-initialized. In Fig. \ref{fig:initialization}, for simplicity, in each partition of lsCNN, only the connections from a neuron in a given feature map to the neurons of the previous layer are shown, as well as the connections of the selected neurons in the first part of lsCNN to their receptive fields in the other parts of such whole model. However, the lsCNN has all the connections of a traditional CNN.

After the initialization, the same training facial images (which were split into patches in the former step) are used to fine tune the weights of the whole lsCNN model, also allowing it to detect some global or more generic features from whole faces, which were not learned locally in the pre-training step.

\section{Experiments, Results and Discussion}

We evaluated the proposed lsCNN architecture on three important face spoofing detection databases: (i) NUAA Imposter Database \cite{bib:NUAA};  (ii) Replay-Attack \cite{bib:Chingovska} dataset; and (iii) CASIA FASD (Face Antispoofing Database)\cite{bib:Zhang}. Subsecs. \ref{sec:NUAA} and \ref{sec:Replay} describe the experiments and the great results obtained, as well as some discussion.
 
\subsection{NUAA Imposter Database}
\label{sec:NUAA}
NUAA Photograph Imposter Database \cite{bib:NUAA} contains grayscale facial photographs (already cropped) obtained from real and fake faces: $3,491$ images for training ($1,743$ from real faces and $1,748$ from printed facial photographs) and $9,123$ images for testing ($3,362$ from real faces and $5,761$ from printed facial images). We performed an initial experiment on this small dataset and, for this, we had to reduced the depth of the lsCNN model, eliminating the third and fourth convolutional and pooling layers due to the small size of the input faces ($64\times64$ pixels - input patches with only $21\times21$ pixels). Given this reduction in depth, for this experiment we augmented the width of the original lsCNN: the first and second convolutional layers presented $90$ and $135$ output feature maps, respectively. The fully-connected layer presented $1,350$ neurons and, following \cite{bib:LeCun}, $5\times5$ kernels (with stride of 2 pixels) were used in the convolutions, given the formed shallow architecture. The first convolutional layer of the lsCNN and of the PatchNets had as input, respectively, $1\times(64\times64)$ and $1\times(21\times21)$ sized feature maps (by working with grayscale images).

The whole lsCNN model was also divided into 9 parts and we initialized all weights of the PatchNets based on random values from a zero-mean normal distribution (with standard deviation of $0.0001$), and normalized the input facial images (before splitting them) by subtracting the mean value of the training set and dividing the values of the pixels by $128$, in order to ensure that most of them would belong to the interval $[-1;1]$. The biases of the neurons were all zero-initialized.

As optimizer, we used the Adam method \cite{bib:Adam}, with the following parameters: $64$ training images per batch, base learning rate of $0.0001$, first momentum of $0.9$ and second momentum of $0.999$. We trained the 9 PatchNets by $2,000$ iterations using the Caffe framework \cite{bib:Caffe}, initialized the whole lsCNN model with their learned weights and biases, and trained the whole CNN for $2,000$ iterations on the whole training faces. For performance comparison, we also assessed a CNN with the same architecture of lsCNN, but traditionally trained, i.e., by initializing all its weights with random values extracted from a normal distribution with zero-mean and standard deviation of $0.0001$ (biases zero-initialized) and training it on the whole faces also by $2,000$ iterations (its convergence point). 

\begin{figure}[h]
\centering
\includegraphics[width=0.4\textwidth]{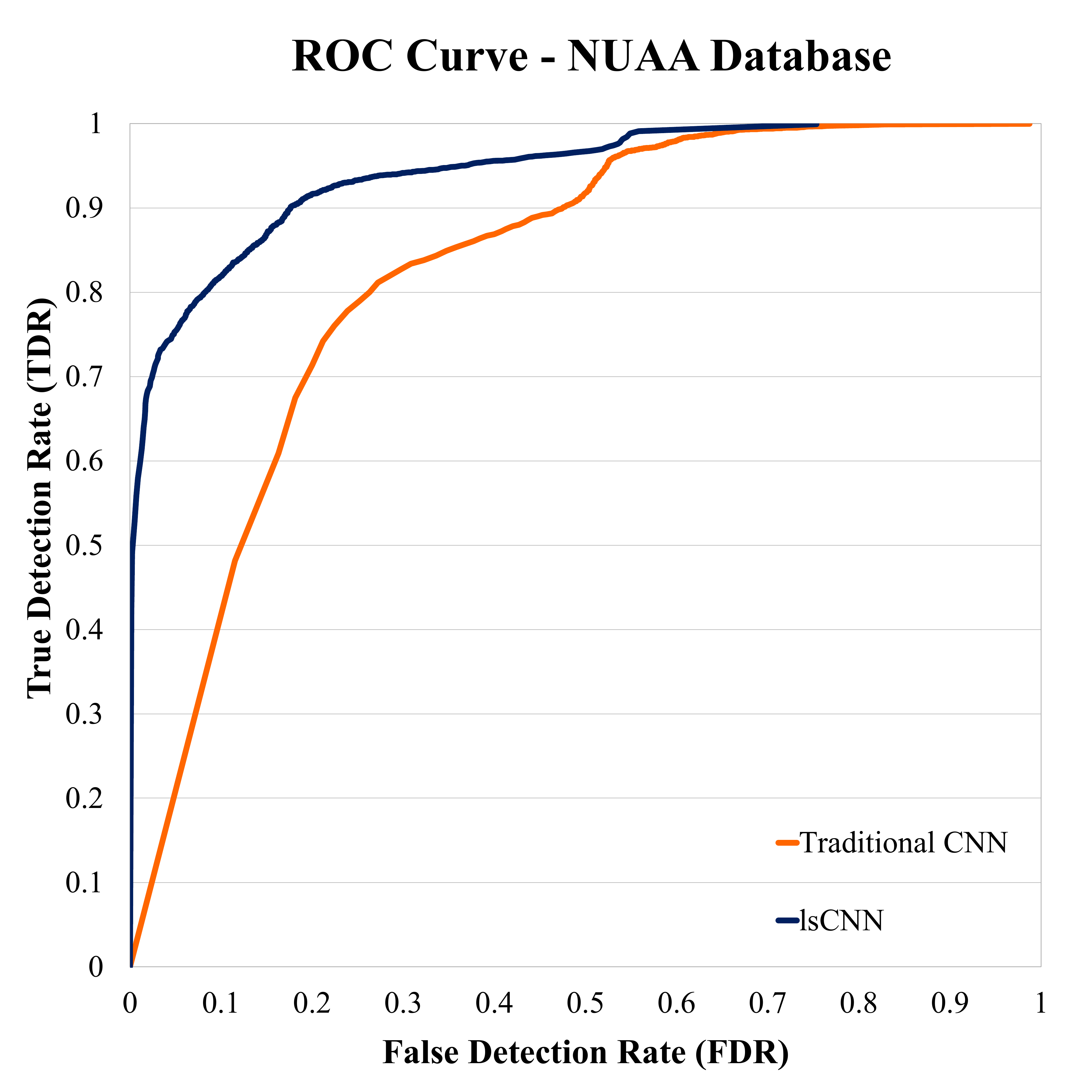}
\caption{ROC curves of the lsCNN and of a CNN with the same architecture, but traditionally trained, i.e., on the whole faces of NUAA \cite{bib:NUAA}, without a local pre-training step. The higher the curve, the better.}
\label{fig:ROCNUAA}
\end{figure}

The goal of this initial experiment was to at first verify the improvement in performance of lsCNN compared with the traditionally trained CNN, given the same amount of training iterations (both $2,000$ iterations). Fig. \ref{fig:ROCNUAA} shows the ROC (Receiver Operating Characteristics) curves of lsCNN and the CNN traditionally trained on whole faces, learning global features. As one can observe, the proposed approach presented a much
better ROC curve than the traditionally trained CNN. Regarding the Equal Error Rate (EER), the lsCNN and the traditionally trained CNN obtained, respectively, 14.10\% and 23.11\%. That is, our proposed approach was again much better than the traditionally trained CNN.

\subsection{Replay-Attack and CASIA Databases}
\label{sec:Replay}

In order to allow a more robust analysis of lsCNN, we performed larger experiments on the Replay-Attack \cite{bib:Chingovska} and CASIA \cite{bib:Zhang} databases. The Replay-Attack dataset contains $360$ videos for training ($60$ videos of real faces and $300$  videos of fake faces), $360$ videos for validation in order to calibrate the threshold of the system used to determine whether a given facial image (extracted from a video frame) is real or fake, and a test set of videos with $80$ videos of real faces and $400$ videos of fake faces. The CASIA \cite{bib:Zhang} dataset presents videos of $50$ subjects, $12$ videos per subject being $3$ of real faces and $9$ of fake faces. The dataset is divided in training set ($20$ subjects, $240$ videos) and test set ($30$ subjects, $360$ videos). There is no validation set explicitly defined for this database.

We detected and cropped the faces in the frames of the videos in both datasets using the robust MTCNN \cite{bib:MTCNN} deep neural network, for an accurate face segmentation. Based on the eyes' landmarks of a face, returned as output by MTCNN, we applied a scale transformation on the respective image  in order to normalize the distance between both eyes to 60 pixels (using the MATLAB algorithm based on interpolation and on the values of the nearest pixels). After detecting and normalizing the face in each frame, we cropped it based on the eyes and capturing the whole facial region (both ears, forehead and chin), with a fixed size of $96\times96$ pixels in RGB color space. Some cropped faces from the Replay-Attack dataset are shown in Fig. \ref{fig:Faces}. In the experiments on both datasets, in order to classify a video, we considered a majority of votes scheme of the faces in its frames. Frames with no face detected by the MTCNN architecture were discarded.

Unlike the experiment with the NUAA dataset, in the experiments with the Replay-Attack and CASIA datasets, we considered the original architecture of lsCNN given the larger facial images obtained. After cropping the faces of all frames of all training videos, an augmentation process on both datasets was performed. In each of them, initially and for each facial image, we generated two new versions of it by increasing or decreasing the values of the R, G, and B channels by $50$. This was done in order to force the network to not rely on brightness for spoofing detection (we did not apply techniques for attenuating the shadows on the faces since they are important to distinguish real faces from 2D fake faces). 

For each of the three versions of each original training facial image, we also applied noise or blur transformations in three levels each (with low magnitudes to not affect the images much), in order to make the neural network also learn smoother features and not rely much on noise. Again we used the MATLAB toolbox for applying blur and Gaussian noise to the images. The blur operation was applied in three levels (using a $2\times2$ Gaussian  filter with standard deviations of $0.1$, $0.5$ and $1.0$), as well as the Gaussian noise (with standard deviations of $0.0005$, $0.00075$ and $0.001$). Such transformations were applied isolatedly, so we obtained, for each of the three initial images from a given face, 6 representations of it. In this sense we augmented our dataset 19 times (original images and $3\times6=18$ transformed images).

For the Replay-Attack dataset we obtained $1,766,031$ training facial images, and for the CASIA dataset, $852,568$ images. Again, we initialized all weights of the smaller PacthNets based on random values from a zero-mean normal distribution (standard deviation of $0.0001$) and normalized each channel of the input facial images by subtracting the mean value of it and diving all the image values by $128$ (before splitting them), in order to ensure that most of them would belong to the interval $[-1;1]$. The biases of the neurons were all zero-initialized. As optimizer, we also used the Adam \cite{bib:Adam} method in both cases, with the same following parameters: $64$ training images per batch, base learning rate of $0.0001$, first momentum of $0.9$ and second momentum of $0.999$. 

In both experiments, we trained the 9 smaller PatchNets for $5,000$ iterations on the facial patches using the Caffe framework \cite{bib:Caffe} and initialized the whole lsCNN model. Then we fine-tuned it over $100,000$ iterations. For the Replay-Attack dataset, the best model was obtained (considering results on the validation set of videos) on iteration $53,600$. For the CNN with the same architecture, traditionally initialized with random values extracted from a normal distribution with zero-mean and standard deviation of $0.0001$ (biases also zero-initialized) and trained on the whole faces, the best model was obtained only on iteration $74,200$ (much later). The results of the proposed approach and of state-of-the-art methods are presented in Tab. \ref{tab:ReplayResults}. For simplicity, we denoted the traditionally trained CNN with the same architecture of lsCNN as ``lsCNN Traditionally Trained".

\begin{table}[h]
\centering
\caption{Results on Replay-Attack \cite{bib:Chingovska} dataset: Equal Error Rate (EER) on the validation dataset and Half-Total Error Rate (HTER) on the test set. Best values are highlighted.}
\begin{tabular}{|c|c|c|} 
\hline
\textbf{Method} & \textbf{EER} & \textbf{HTER}  \\ 
\hline \hline
Efficient Fine-Tuned VGG-Face \cite{bib:CIARP2017} & --- & 16.62 \\ \hline
Patch Based Handcrafted Approach \cite{bib:Ak} & --- & 5.0 \\ \hline
Whole Fine-Tuned VGG-Face \cite{bib:Lotufo} & --- & \textbf{1.20} \\ \hline 
Fine-Tuned VGG Face\cite{bib:Li} & 8.40 & 4.30 \\ \hline 

Li et al. \cite{bib:Li} & 2.90 & 6.10 \\ \hline 
Random Patches Based CNN \cite{bib:Atoum} & 2.50 & \textbf{1.25} \\ \hline 
Boulkenafet et al. \cite{bib:ref1} & 0.40 & 2.90 \\ \hline
lsCNN Traditionally Trained & \textbf{0.33} & 1.75 \\ \hline
lsCNN & \textbf{0.33} & 2.50 \\ \hline 
\end{tabular}
\label{tab:ReplayResults}
\end{table}

As one can observe, besides obtaining the best EER, lsCNN presented a great HTER, much better than expensive methods, such as \cite{bib:Lotufo}, which work with extremely complex and large CNNs, such as VGG-Face \cite{vggface}. Despite  obtaining a worse HTER result than the traditionally trained neural network, lsCNN obtained the presented results much faster (in a much earlier iteration of the training), as mentioned. 

Regarding the CASIA experiment, the best model for lsCNN was obtained on iteration $9,800$, while the best model for the traditionally trained CNN was obtained on iteration $80,900$. In order to compare the performances of such methods with state-of-the-art approaches, we measured the EER, since this dataset presents a predefined test dataset.  Tab. \ref{tab:ResultsCASIA} shows the results.

\begin{table}[h]
\centering
\caption{Results in the CASIA \cite{bib:Zhang} dataset of the proposed network architecture (lsCNN) and other state-of-the-art methods. The best values are highlighted.}
\label{tab:ResultsCASIA}

\begin{tabular}{| c | c |}
\hline
\textbf{Method} & \textbf{EER} \\ 
\hline\hline 
Fine-tuned VGG-Face \cite{bib:Li}& 5.20 \\ \hline 
LSTM-CNN  \cite{bib:Xu} & 5.17 \\ \hline
Yang et al. \cite{bib:ref3} & 4.92 \\ \hline

Patch Based Handcrafted Approach \cite{bib:Ak} & 4.65 \\ \hline 
Li et al. \cite{bib:Li}&4.50\\ \hline 
Random Patches Based CNN \cite{bib:Atoum}&\textbf{4.44} \\ \hline 

lsCNN Traditionally Trained&\textbf{4.44} \\ \hline 

lsCNN&\textbf{4.44} \\ \hline 

\end{tabular}
\end{table}

As one can observe, lsCNN obtained the best EER on the CASIA dataset, as well as the traditionally trained CNN and the work of \cite{bib:Atoum}, better than approaches that require complex and expensive architectures. Besides, when compared with the traditionally trained CNN, lsCNN training was much faster (lsCNN obtained its best performance on iteration $9,800$ against iteration $80,900$ for the lsCNN architecture traditionally trained).

\section{Conclusion}
Face spoofing detection is a critical task nowadays, given
the widespread usage of face recognition systems and the development by criminals of attack techniques to simulate faces of legal users. Traditional face recognition systems can be easily circumvented  with common printed facial photographs, available, nowadays, in social medias and networks.


Despite the fact that face detection and recognition methods take into account the different regions of human face for such tasks, to the best of our knowledge, no technique used deep local spoofing cues for attack detection so far, as we propose. Experimental results show a high increase in the performance of the proposed CNN architecture, lsCNN, when initialized based on a local pre-training step (on the main facial regions). The lsCNN obtained state-of-the-art results on the evaluated datasets with a quite compact model, also being much more efficient than benchmark CNNs, such as VGG-Face, which is highly used for attack detection through transfer learning. 

The proposed training approach can also be applied for training other CNN models, including larger architectures, in order to improve their performances in spoofing detection as well as their efficiency during learning even more. 

\section*{Acknowledgements}
The authors are grateful to FAPESP (grants \#2014/12236-1, \#2017/05522-6 and \#2016/19403-6), CAPES (grant \#88881.132647/2016-01), to Dr. Anil K. Jain for the doctoral exchange period, to NVIDIA, and to Banco do Brasil.

\bibliographystyle{IEEEtran}

%
%


\end{document}